\newcommand{\citet}[1]{\citeauthor{#1}~\shortcite{#1}}
\newcommand{\citep}{\cite}

\def\year{2019}\relax
\documentclass[letterpaper]{article} 
\usepackage{aaai19}  
\usepackage{times}  
\usepackage{helvet}  
\usepackage{courier}  
\usepackage[hyphens]{url}  
\usepackage{nameref}
\usepackage{color}
\usepackage{graphicx}  
\usepackage{lipsum}
\usepackage{amsmath}
\usepackage{amssymb}
\usepackage[ruled]{algorithm2e}
\usepackage{booktabs} 
\usepackage{tabularx}
\usepackage{threeparttable}
\usepackage{subcaption} 
\usepackage{placeins}
\usepackage{wasysym}
\usepackage{mathtools}
\begin{document}
%
 \author{Muhammad Raza Khan,
	Joshua E. Blumenstock\\
	{University of California, Berkeley}\\
	mraza@bekeley.edu,
	jblumenstock@berkeley.edu 
	}
\title{Multi-GCN: Graph Convolutional Networks for Multi-View Networks,\\ with Applications to Global Poverty}


\maketitle
\begin{abstract}
With the rapid expansion of mobile phone networks in developing countries, large-scale graph machine learning has gained sudden relevance in the study of global poverty. Recent applications range from humanitarian response and poverty estimation to urban planning and epidemic containment. Yet the vast majority of computational tools and algorithms used in these applications do not account for the multi-view nature of social networks: people are related in myriad ways, but most graph learning models treat relations as binary.  In this paper, we develop a graph-based convolutional network for learning on multi-view networks. We show that this method outperforms state-of-the-art semi-supervised learning algorithms on three different prediction tasks using mobile phone datasets from three different developing countries. We also show that, while designed specifically for use in poverty research, the algorithm also outperforms existing benchmarks on a broader set of learning tasks on multi-view networks, including node labelling in citation networks.
\end{abstract}

\section{Introduction}

Over the past several years, large-scale graph machine learning has gained increasing relevance in the domain of international poverty research~\cite{blumenstock_fighting_2016}. Driven largely by the expansion of mobile phone networks throughout developing countries -- roughly 95\% of the world population now has mobile phone coverage~\cite{gsma_unlocking_2016}
-- vast quantities of network data are constantly being generated by people living in even extremely poor and marginalized communities. Recent work has shown how such data can be used to inform critical policy decisions, including the measurement of living conditions \cite{blumenstock2015predicting}, the spread of infectious diseases \cite{wesolowski2015dengue}, and the management of humanitarian crises \cite{lu2012haitiearthquake}. Private companies are also taking advantage of this new source of data, for instance by using data from mobile phones to generate credit scores that can expand credit to millions of people historically shut out of the formal banking ecosystem \cite{francis2017digital}.

However, a critical constraint to the use of these data in settings related to economic development is the lack of scalable algorithms for performing prediction tasks on sparse multi-view networks. Multi-view networks (also referred to as multiplex and multi-modal networks), are networks in which nodes can be related in multiple ways, and are the natural abstraction for mobile phone networks, where different individuals have different types of relationships and can interact using different modalities (such as phone calls, text messages, money transfers, and app-based activity). Yet, the vast majority of applied research using mobile phone data --- in developing and developed countries alike --- ignores the multi-view nature of phone networks.

This paper develops a novel approach for learning on multi-view networks, which bridges two different strands in the research literature. The first strand involves methods for efficient analysis of multi-view networks; the second explores algorithms for semi-supervised graph learning (see Related Work, below). The method we develop provides an efficient approach for applying convolutional neural networks to multi-view graph-structured data. 
We benchmark this new method, which we call Multi-GCN (short for Multi-View Graph Convolutional Networks), on three different mobile network datasets, on three different prediction tasks relevant to the international development community: (1) predicting the adoption of a new ``financial inclusion'' technology in a West African country; (2) predicting whether an individual is living below the poverty line in an East African country; (3) predicting the gender of mobile phone subscribers in a South Asian country. In all cases, we find that Multi-GCN outperforms state-of-the-art benchmarks, including standard Graph Convolutional Networks \cite{kipf2016semi}, Node2Vec \cite{node2vec}, Deepwalk \cite{deepwalk}, and LINE \cite{tang2015line}. 

While designed specifically with the developing-country context in mind (where the sparsity and multi-view properties of networks are very salient), we show that Multi-GCN can be more generally applied to a wide range of problems involving multi-view networks. Indeed, most real-world networks are multi-view, including the network data most frequently used by AI researchers (e.g., data from Twitter, Amazon, Netflix, etc.). Our second set of results shows that Multi-GCN can improve upon state-of-the-art algorithms not just in poverty-related contexts, but also in traditional classification problems. In particular, we show that Multi-GCN outperforms competing algorithms on citation labeling tasks (using benchmark datasets from Citeseer and Cora) that have been studied extensively in prior work.

\section{Related Work}
\begin{figure*}[!h]
	\centering
	\begin{subfigure}{.14\textwidth}
		\centering
		\includegraphics[width=.9\linewidth]{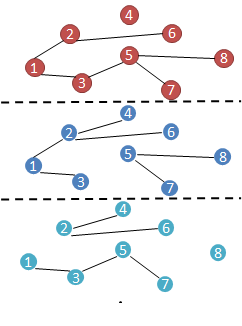}
		\begin{minipage}{0.9\textwidth} 
			{\footnotesize \textit{Multi-view graph}
				G=((V,E1),(V,E2),(V,E3))\par}
		\end{minipage}
	\end{subfigure}%
	$\xRightarrow[\text{Fusion}]{\text{Multiview}}$%
	\begin{subfigure}{.2\textwidth}
		\centering
		\includegraphics[width=.9\linewidth]{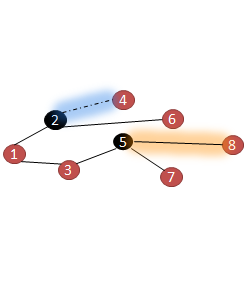}
		\begin{minipage}{0.9\textwidth} 
			{\footnotesize \textit{Merged graph}\\
				(Centroids in black,\\
				salient edges in blue,\\
				other edges in orange }
		\end{minipage}
	\end{subfigure}
	$\xRightarrow[\text{Ranking}]{\text{Manifold}}$%
	\begin{subfigure}{.2\textwidth}
		\centering
		\includegraphics[width=1\linewidth]{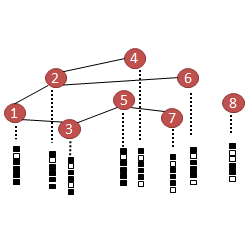}
		\begin{minipage}{0.9\textwidth} 
			{\footnotesize \textit{Rank-augmented graph and node features}\\
				(after adding salient edges, pruning others)}
		\end{minipage}
	\end{subfigure}
	$\xRightarrow[\text{to GCN}]{\text{Input}}$%
	\begin{subfigure}{.26\textwidth}
		\centering
		\includegraphics[width=0.9\linewidth]{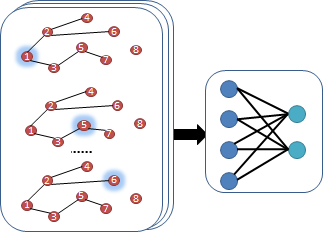}
		\begin{minipage}{0.8\textwidth} 
			{\centering\footnotesize \textit{Graph Convolution Network}
				Hidden layers$\rightarrow$Dense layers}
		\end{minipage}
	\end{subfigure}
	\caption{Overview of the Multi-view Graph Convolutional Network (Multi-GCN)}\label{fig:workflow}

\end{figure*}
\subsection{Technical Related Work}
\label{sec:trw}
Our goal is to develop an efficient method for node-level transductive semi-supervised learning over multi-view graphs. Here, we begin with a general overview of semi-supervised learning, then focus on various approaches to graph-based semi-supervised learning, and finally discuss related work on multi-view networks.

\subsubsection{Graph-Based Semi-Supervised Learning}

One of the biggest issue with applying supervised learning algorithms in a developing country  is that it is often costly to collect labels for training. For instance, when using mobile phone data to predict the wealth of subscribers, \citet{blumenstock2015predicting} manually conducted a survey of roughly 1,000 subscribers. Semi-supervised learning tries to solve this problem by using unlabeled data along with the labeled data to train better classifiers (see \cite{zhu05survey} for a survey). Our focus is on transductive semi-supervised learning, which assumes that all the unlabeled data is available at the training time and does not attempt to generalize to data unseen during training. 

Graph-based semi-supervised learning (GSSL) is a popular approach for semi-supervised learning that treats labeled and unlabeled instances as graph vertices, and relationships between instances as edges \cite{liu2012robust}.
GSSL algorithms try to learn a classifier that is consistent with the labeled data while making sure that the prediction for similar nodes is also similar. This is achieved by minimizing a loss function with two factors: a) supervised loss over the labeled instances, and b) a graph-based regularization term.  Different GSSL algorithms use different functions for graph regularization. 
Label propagation-based approaches, for instance, use a constrained label lookup function (e.g., \citet{zhou2004learning}). Related, kernel-based approaches parameterize regularization term in the Reproducing Kernel Hilbert Space (RKHS). 

\subsubsection{Learning Over Graphs}
The success of word embedding algorithms like Word2Vec \cite{mikolov2013distributed} has inspired similar algorithms for graphs. For instance, DeepWalk \cite{deepwalk} learns embeddings by predicting the neighborhood of nodes based on random walks over the graphs, while LINE \cite{tang2015line} and Node2vec \cite{node2vec} allow for advanced sampling schemes. 
More recently, neural network-based approaches have been proposed to perform learning over graphs. These have been extended to the task of semi-supervised learning \cite{bruna2013spectral,defferrard2016convolutional}, including recent work by \citet{kipf2016semi} that proposes a Graph Convolutional Network (GCN), which we take as a starting point for our approach.

\subsubsection{Learning Over Multi-View Graphs}

The key distinction between our approach and prior work is our desire to handle graphs with multiple views, i.e., graphs where vertices can be connected in more than one way.  In recent years, many different algorithms have been proposed for learning on multi-view graphs. These algorithms can be broadly divided into three main categories: 1) co-training algorithms, 2) learning with multiple kernels, and 3) subspace learning (See \citet{xu2013survey} for a survey).
Recent work by \citet{dong2014subspace} show that subspace approaches --- which find a latent subspace shared by multiple views --- perform well relative to co-training and kernelized approaches on a range of tasks. We therefore focus our attention on integrating subspace learning approaches with recent innovations in graph convolutional networks.

\subsubsection{Comparison with existing work}

Our main contribution is to propose an efficient method for adapting GSSL to multi-view contexts. Existing approaches to GSSL cannot be readily implemented on such data; those algorithms that do handle multiple views generally treat views and vertices equally.
We show that current ``state of the art'' methods like Graph Convolutional Networks \cite{kipf2016semi} can be enhanced by augmenting the input graph using subspace analysis over Grassman manifolds. \citet{farseev2017cross} have demonstrated that subspace merging approach can be quite accurate for the problem of cross-domain recommendation which is different from our experimental settings and context as described in the section \ref{Experiments}.

\subsection{Empirical Related Work} 

Our experimental results focus on three prediction tasks of relevance to the international development community:

\subsubsection{Predicting poverty.} A large number of humanitarian applications --- from poverty targeting to program monitoring --- require accurate estimates of the welfare for beneficiary populations. Recently, several papers have shown how digital trace data can be used to estimate the socioeconomic status of individuals, households, and villages.  For instance, \citet{jean2016combining} show that daytime satellite imagery can be used to estimate village wealth; \citet{quercia2012tracking} find that Twitter data can be used to estimate levels of deprivation, and \citeauthor{blumenstock2014} (2015) shows that mobile phone metadata can be used to estimate the welfare of individuals and regions.

\subsubsection{Product adoption.} We focus on the adoption of ``mobile money'', a suite of phone-based financial services that are designed to promote financial inclusion among those traditionally shut out of the formal banking ecosystem \cite{suri}. 
Within this literature, our work relates most closely to \citeauthor{kdd_2016_mrk} (2016), who analyze the predictors of mobile money adoption in three different developing countries. 

\subsubsection{Gender prediction.} Gender equality and women's empowerment are one of the Sustainable Development Goals, and recent work explores how digital trace data can be used to assess progress toward this goal \cite{fatehkia}. \citet{mislove2011understanding} and \citet{frias2010gender} show that gender can be predicted from social media and mobile phone data.

Broadly, these prior studies demonstrate a proof of concept: that digital trace data can be used to predict the characteristics and outcomes of individuals. However, such analysis rely on off-the-shelf algorithms that rarely, if ever, account for the multi-view nature of real-world social networks. This paper shows that a simple approach to multi-view learning can yield substantial improvements on these real-world prediction tasks.

\section{\textit{Multi-GCN}: Multi-View Graph Convolutional Networks}\label{sec:methods}
Our approach to semi-supervised learning on multi-view graphs integrates three steps, depicted in Figure~\ref{fig:workflow}. First, we use methods from subspace analysis to efficiently merge multiple views of the same graph. Second, we use a manifold ranking procedure to identify the most informative sub-components of the graph and to prune the graph upon which learning is performed. Finally, we apply a convolutional neural network, adapted to graph-structured data, to allow for semi-supervised node classification. 
\subsection{Merging Subspace Representations}\label{subsec:methods_subspace}
Given an undirected multilayer graph with M layers $G={G_i}_{i=1}^M$ such that each layer $G_i$ has the same vertex set $V$ but same or different edges set $E_i$, we first calculate the graph Laplacian for each of the individual layers.
If $D_i$ and $W_i$ represent the degree matrix and the adjacency matrix for the $i^{th}$ view of the graph, then the normalized graph Laplacian is defined as  
\begin{eqnarray}
\label{eq:2}
\begin{aligned}
L_i=D_i^{-1/2}(D_i-W_i)D_i^{-1/2}
\end{aligned}
\end{eqnarray}

Given the graph Laplacian $L_i$ for each layer of the graph, we calculate the spectral embedding matrix $U_i$ through trace minimization:
\begin{eqnarray}
\label{eq:3}
\begin{aligned}
\min_{U_i \in \mathbb{R}^{n*k}} tr(U_i' L_i U_i), & &\textnormal{s.t. } U_i'U_i=1
\end{aligned}
\end{eqnarray}

This trace minimization problem can be solved by the Rayleigh-Ritz theorem. The solution $U_i$ contains the first $k$ eigenvectors corresponding to the $k$ smallest eigenvalues of $L_i$. The spectral embedding embeds nodes of the original graph to a low dimensional spectral domain (See \citet{von2007tutorial} for details).

A Grassman manifold $\mathcal{G}(k,n)$ can be considered as a set of $k$-dimensional linear subspaces in $\mathbb{R}^n$ where each unique subspace is mapped to a unique point on the manifold. Each point on the manifold can be represented by an orthonormal matrix $Y \in \mathbb{R}^{n*k}$ whose columns span the corresponding k-dimensional subspace in  $\mathbb{R}^{n*k}$ and the distance between the subspaces can be calculated as a set of principal angles $\{\theta_i\}_{i=1}^k$ between these subspaces. \citet{dong2014subspace} show that the projection distance between two subspaces $Y_1$ and $Y_2$ can be represented as a separate trace minimization problem:
\begin{eqnarray}
\label{eq:4}
\begin{aligned}
d_{proj}^2(Y_1,Y_2)=\sum_{i=1}^{k}\sin^2\theta_i=k-tr(Y_1Y_1'Y_2Y_2')
\end{aligned}
\end{eqnarray}
where, based on Eq. \ref{eq:4}, the projection distance between the target representative subspace $U$ and the individual subspaces ${U_i}_{i=1}^M$ can be calculated as:
\begin{eqnarray}
\label{eq:5}
\begin{aligned}
d_{proj}^2(U,\{U_i\}_{i=1}^M) & = &\sum_{i=1}^{M}d_{proj}^2(U,U_i)\\
& = &kM-\sum_{i=1}^{M}tr(UU'U_iU_i')
\end{aligned}
\end{eqnarray}
Minimization of Eq. \ref{eq:5} ensures that individual subspaces are close to the final representative subspace $U$. 

Finally, to ensure that the original vertex connectivity in each graph layer is preserved, we include a separate term that minimizes the quadratic-form Laplacian (evaluated on the columns of U):
\begin{eqnarray}
\label{eq:6}
\begin{aligned}
\min_{U \in \mathbb{R}^{n*k}} \sum_{i=1}^{M} tr(U' L_i U)+\alpha_i[kM-tr(UU'U_iU_i')],\\ \textnormal{s.t. } U_i'U=1
\end{aligned}
\end{eqnarray}
In Eq \ref{eq:6}, $\alpha$ is the regularization parameter that balances the trade-off between the two terms in the objective function. Rearranging Eq. \ref{eq:6} and ignoring the constant terms yields
\begin{eqnarray}
\label{eq:7}
\begin{aligned}
\min_{U \in \mathbb{R}^{n*k}} tr[U'(\sum_{i=1}^{M}  L_i -\sum_{i=1}^{M}\alpha_iU_iU_i')U],\\ 
\end{aligned}
\end{eqnarray}
As before, the Rayleigh-Ritz theorem can be used to solve Eq \ref{eq:6}. The solution is given by the fist $k$ eigenvectors of the modified Laplacian:
\begin{eqnarray}
\label{eq:8}
\begin{aligned}
L_{mod}=\sum_{i=1}^{M}L_i -\sum_{i=1}^{M}\alpha_iU_iU_i'
\end{aligned}
\end{eqnarray}
\subsection{Graph-Based Manifold Ranking}\label{subsec:methodsranking}

Though the modified Laplacian calculated above can be fed directly to the downstream graph convolutional networks, model performance can be increased by ranking the nodes in the manifold based on their saliency with respect to some critical nodes \cite{zhou2004ranking}. To rank points on the manifold, we use the closed form function,
\begin{eqnarray}
\label{eq:9}
\begin{aligned}
f^*=(I-\beta*L_{mod})^{-1}q
\end{aligned}
\end{eqnarray}
Here, $I$ represents the identity matrix, $L_{mod}$ is the normalized Laplacian as calculated in Eq. \ref{eq:8}, and $\beta$ is the regularization parameter.  Given a vector $q$ containing the indices of the query nodes, Eq. \ref{eq:9} calculates the saliency of the other nodes with respect to the query nodes; the saliency of these nodes can then be used to add or prune edges from the induced underlying graph. The use of manifold-based ranking suits our approach as the modified Laplacian representing merged subspaces can be used directly for saliency detection. The query nodes can be selected as the centroids determined by any clustering algorithm over the manifold.

The algorithm for the subspace merging and subsequent manifold ranking is shown in Algorithm \ref{algorithm}. The time complexity of Algorithm \ref{algorithm} for a graph with $M$ layers with $N$ users per layer is $O(MN^3+MN^2K+N^2C^2+tN)$ where $K$ represents the number of eigenvectors to be calculated and $C$ is the number of centroids $O(MN^3)$ is the cost of computing Laplacians and Eigenvector matrix for all the $M$ layers ; $O(MN^2K)$ is the cost of computing modified Laplacian; $O(N^2C^2)$ is the cost of computing $C$ clusters using k-means clustering; $O(tN)$ is the cost of manifold ranking. using the iterative version described by \cite{zhou2004ranking}.

\setlength{\textfloatsep}{10pt}
\begin{algorithm}[!htb]
	
	\SetAlgoLined 
	\caption{Fusion of multiple views of a graph\label{algorithm}}
	\BlankLine
	\BlankLine
	
	\KwIn{\{$A_i\}_{i=1}^M$: $n \times n$ adjacency matrices of individual graph layers $\{G_i\}_{i=1}^M$, with $G_1$ being the most informative layer}
	\KwIn{$\{\alpha_i\}_{i=1}^M$,regularization parameters per subspace to be merged}
	\KwIn{$K$, salient query points}
	\KwIn{$Y$, number of salient edges per centroid to add}
	\KwIn{$Z$, number of non-salient edges per centroid to prune}
	\KwIn{$\beta$, manifold ranking regularizer}
	\BlankLine
	\KwOut{$L_{mod}$:Merged Laplacian,$A_{mod}$: Merged Adjacency matrix,  $E_{s}$:Salient Edges, $E_{ns}$: Non salient edges}
	\hrulefill
	
	\textbf{Step 1:} Compute normalized Laplacian matrix $L_i$ for each layer of the graph\\
	\textbf{Step 2:} Compute subspace representation $U_i$ for each layer of the graph\\
	\textbf{Step 3:} Compute the modified Laplacian matrix $L_{mod}=\sum_{i=1}^{M} L_i -\sum_{i=1}^{M}\alpha_iU_iU_i'$ \\
	\textbf{Step 4:} Perform clustering on the modified Laplacian to identify $K$ salient points i.e. centroids $\{q_i\}_{i=1}^K$\\
	\textbf{Step 5:} For each of the centroid rank other edges on the manifold $f^*=(\mathcal{I}-\beta*L_{mod})^{-1}q$\\
	\textbf{Step 6:} For each centroid $q_i$ add $Y$ salient edges to the $E_{s}$ and $Z$ non-salient edges to the $E_{ns}$\\
	
	\textbf{Step 7:} Add $E_{s}$ to $A_1$ to form $A_{mod}$\\
	\textbf{Step 8:} Remove  $E_{ns}$ from $A_{mod}$\\
	
\end{algorithm}

\begin{table*}[!t]\centering
	\begin{threeparttable}
		\renewcommand{\arraystretch}{1.3}
		\begin{tabular}{l*{2}{cccccc}}
			\addlinespace
			\toprule
			\textbf{Dataset} & \textbf{Data Type} & \textbf{Nodes} & \textbf{Edges}& \textbf{Edges}&\textbf{Classes} & \textbf{Features} & \textbf{Label Rate} \\
			&&& \textbf{(view 1)}& \textbf{(view 2)}&&& \\
			\midrule
			Product Adoption & Phone logs (West Africa) & 17,000  & 23,032& 18,371 & 2 & 132 & 0.002\\
			Poverty Prediction & Phone logs (East Africa) & 422  & 544& 1,799 & 2 & 1,709 & 0.094\\
			Gender Prediction & Phone logs (South Asia) & 958 & 992 & 978 & 2 & 821 & 0.042\\
			\midrule
			Citeseer & Citation network & 3,327 & 4,732 & 3,492 & 6 & 3,703 & 0.036\\
			Cora& Citation network & 2,708  & 5,429& 2,846 & 7 & 1,433 & 0.052\\
			\bottomrule
		\end{tabular}
		\caption{\textbf{Summary statistics.} \normalfont{The Label Rate indicates the fraction of instances that are labeled.} \label{tab:sumstats}}
	\end{threeparttable}
\end{table*}

\begin{figure*}[!htb]
	\centering
	\begin{subfigure}{0.33\textwidth}
		\centering
		\includegraphics[width=\linewidth]{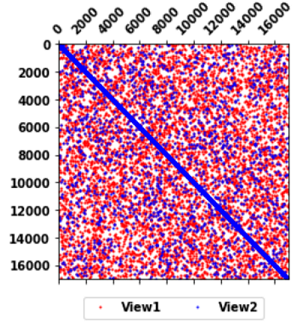}
		\caption{Product Adoption}
		\label{fig:productadoption_spy}
	\end{subfigure}
	\hfill
	\begin{subfigure}{0.33\textwidth}
		\centering
		\includegraphics[width=\linewidth]{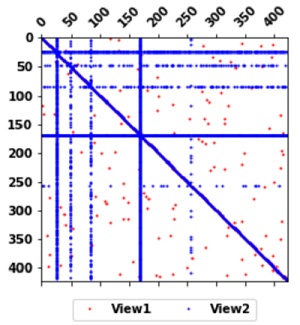}
		\caption{Wealth Prediction }
		\label{fig:wealth_spy}
	\end{subfigure}
	\hfill
	\begin{subfigure}{0.33\linewidth}
		\centering
		\includegraphics[width=\linewidth]{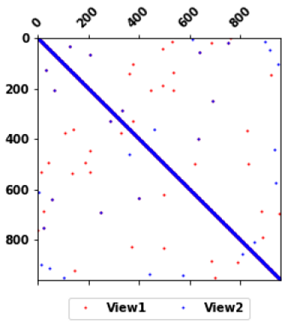}	
		\caption{Gender Prediction }
		\label{fig:gender_spy}
	\end{subfigure}
	\caption{\textbf{Mobile phone spy plots.} \normalfont{Dots indicate that two individuals have communicated by voice (red) or SMS (blue).} \label{fig:spy}}
\end{figure*}

\subsection{Graph Convolution Networks}\label{subsec:methods_gcn}
The application of convolutional neural networks to irregular or non-Euclidean grids, such as graphs, is based on the fact that convolutions are multiplications in the Fourier domain, which implies that graph convolutions can be expressed as the multiplication of a signal $x \in \mathbb{R}^N$ with a filter $g(\theta)$ (see \citet{bruna2013spectral}):
\begin{eqnarray}
\label{eq:10}
\begin{aligned}
g_{\theta}*x=g_{\theta}(L)x=Ug_{\theta}U^Tx
\end{aligned}
\end{eqnarray}
Here, $U$ represents the eigen-decomposition of the normalized graph Laplacian $L=I-D^{-1/2}AD^{-1/2}$ and  $I$, $D$, $A$ represent the identity, degree and the adjacency matrix, respectively. Graph convolutions can be further expressed in terms of Chebyshev polynomials as
\begin{eqnarray}
\label{eq:11}
\begin{aligned}
g_{\theta'}*x=\sum_{k=0}^{K}\theta_k'T_k(\tilde{L})x
\end{aligned}
\end{eqnarray}
where $\tilde{L}$ is the rescaled Laplacian, $T_k$ represents the Chebyshev polynomials, and $\theta'$ represents the vector of Chebyshev coefficients. Following \citet{kipf2016semi}, by approximating the maximum value of the largest eigenvalue and constraining the number of free parameters, the convolution operation can be represented as 
\begin{eqnarray}
\label{eq:12}
\begin{aligned}
g_{\theta}*x=\theta(I+\tilde{D}^{-1/2}\tilde{A}\tilde{D}^{-1/2})x
\end{aligned}
\end{eqnarray}
where $\tilde{A}=A+I$ and $\tilde{D}=\sum \tilde{A} $ are the renormalized versions of $A$ and $D$. This renormalization avoids numerical instabilities resulting from exploding/vanishing gradients \cite{defferrard2016convolutional}.

The modified graph ($A_{mod}$ in Algorithm~\ref{algorithm}) resulting from the merger of Laplacians using the subspace analysis and manifold ranking can be fed directly into the graph convolution networks defined above. The forward propagation model for a two layer network can then be represented as 
\begin{eqnarray}
\label{eq:13}
\begin{aligned}
Z=F(X,A)=softmax\large(\hat{A}\thinspace  ReLU(\hat{A}XW^0)W^{1}\large)
\end{aligned}
\end{eqnarray}
Here, $\hat{A}=\tilde{D}^{-1/2}\tilde{A}\tilde{D}^{-1/2}$ is calculated as a preprocessing step before giving the input to the neural network. 
$W^0$ and $W^1$ represent the input-to-hidden-layer and hidden-layer-to-output weight matrices for a two layer neural network, and can be trained using gradient descent. ReLU and Softmax represent the activation functions in the hidden and output layers. 


\section{Experiments and Data}\label{Experiments}
\subsection{Datasets}\label{sec:datasets}
Our first set of experiments test Multi-GCN on three prediction tasks relevant to international development. Each one uses a different dataset of mobile phone Call Detail Records (CDR), obtained from three different developing countries with GDP per capita less than \$1,600 USD. These datasets contain detailed metadata on all communication events (calls, messages) that occur on the mobile phone network. Each CDR dataset contains multiple possible relationships between nodes (views); we extract one view corresponding to phone calls between users, and another corresponding to text messages. We separately construct a large set of features of each user (such as total call volume and degree centrality), using the combinatoric approach described in \citet{kdd_2016_mrk}.

Table~\ref{tab:sumstats} presents summary statistics for each of these datasets. The connections and sparsity of each network are shown in Figure~ \ref{fig:spy}. These spy plots help visualize the structure of the adjacency matrices for each graph view, where a dot indicates that an edge exists between those two individuals on the corresponding view.

\begin{table*}[!ht]\centering
	\begin{threeparttable}
		\renewcommand{\arraystretch}{1.2}
		\begin{tabular}{l*{1}{ccc}}
			\addlinespace
			\toprule
			
			\midrule 
			\textbf{Method} & \textbf{Product Adoption} & \textbf{Poverty Prediction}& \textbf{Gender Prediction}\\
			\midrule
			DeepWalk (first view)  & 56.43$\pm$0.187& 51.91$\pm$0.62 & 53.18$\pm$ 0.55 \\
			DeepWalk (second view) & 51.97$\pm$0.112& 50.34$\pm$0.36  & 50.84$\pm$0.64\\
			DeepWalk (view union) & 56.81$\pm$ 0.114& 50.87$\pm$0.95  & 52.34$\pm$0.50\\
			Node2vec (first view)  & 53.87$\pm$0.20& 52.26$\pm$0.58 & 50.12$\pm$ 0.40 \\
			Node2vec (second view) & 50.50$\pm$0.11& 49.70$\pm$0.23  & 51.68$\pm$0.40\\
			Node2vec (view union) & 54.50$\pm$0.11& 50.52$\pm$0.63  & 51.64$\pm$0.53\\
			LINE (first view)  & 51.11$\pm$0.01& 50.15$\pm$0.02 & 51.56$\pm$ 0.001 \\
			LINE (second view) & 50.83$\pm$0.01& 52.29$\pm$0.001  & 50.00$\pm$0.001\\
			LINE (view union) & 56.26$\pm$0.003& 50.18$\pm$0.001  & 51.33$\pm$0.002\\
			
			GCN (first view)  & 70.74$\pm$2.2& 55.19$\pm$2.33 & 63.97$\pm$ 1.29 \\
			GCN (second view) & 71.40$\pm$1.81& 50.06$\pm$0.81  & 63.01$\pm$0.013\\
			GCN (view union) & 71.90$\pm$0.9& 50.22$\pm$0.56  & 63.90$\pm$1.32\\
			\textbf{Multi-GCN (this paper)} & \textbf{73.47$\pm$0.91}& \textbf{59.23$\pm$0.20}  & \textbf{66.34$\pm$ 1.03}\\
			\midrule
			
			\bottomrule
		\end{tabular}
		\caption{\textbf{Classification accuracy on mobile phone data}. \normalfont{Numbers indicate mean classification accuracy (percentage) and standard error over 10 randomly selected dataset splits of equal size.}\label{tab:results_random}}
	\end{threeparttable}
\end{table*}

\subsubsection*{\textbf{Product adoption dataset}}~\\
The first dataset that we use is a sample of a dataset of mobile phone activity from a West African country.  Here, the classification of interest is whether or not the user eventually adopts a new financial inclusion product. There are two possible classifications: (1) Did not adopt; (2) Adopted and used the product. Following the experimental setup described in \citet{kipf2016semi}, we randomly selected 20 users from each category (40 total) for the training dataset; the validation and the testing dataset consist of 500 and 1000 randomly selected users, respectively.

\subsubsection*{\textbf{Poverty prediction dataset}}~\\
The wealth prediction dataset consists of several thousand transactions of different mobile phone users from an East African country. We attempt to classify users as poor or non-poor, where labels were obtained by \citet{blumenstock2015predicting} through a small set of phone surveys that were conducted with mobile phone subscribers. Again, we randomly selected 20 users from each category as the training dataset, while the size of the validation dataset and the testing dataset is 100 and 200 respectively. 

\subsubsection*{\textbf{Gender prediction dataset}}~\\
The gender prediction dataset originates from a developing country in South Asia. Here, the classification task is to predict the gender of the mobile phone users, where gender labels are provided by the operator for a small number of labeled instances. We randomly select 20 users from each category for training; the size of the validation and the testing datasets are 100 and 800, respectively.

\subsubsection*{\textbf{Citation classification datasets}}~\\
A final set of experiments replicates the experimental design of \citet{kipf2016semi} to test Multi-GCN on more standard node labelling tasks. 
In these datasets, nodes are documents and the first view corresponds to the citation links between the research papers. We construct the
second view from the textual similarity of the papers. Specifically, 
if the normalized cosine similarity between documents is greater than 0.8, then we create an edge in the second view of the citation network. 

\subsection{Experimental setup}
In general, our goal is to correctly classify nodes in a network, where only a very small fraction of nodes are labeled. In the experiments, we start from a small sample of labeled nodes and test the ability of Multi-GCN, as well as several state-of-the-art algorithms, to correctly classify unlabeled nodes in the validation and testing sets. We use three popular node embedding algorithms (Node2vec, Deepwalk, and LINE) as a first set of baselines. In addition, we provide three baselines based on graph convolutional networks \cite{kipf2016semi}. The first two, \textit{GCN (first view)} and \textit{GCN (second view)}, apply GCN over the two respective adjacency matrices from phone and text message activity. The third, \textit{GCN (view union)}, operates on the union of the adjacency matrices of the first view and the second view. In each GCN baseline, the node features are constructed from the adjacency matrix of the first view.

After merging different views, we rank the interaction between nodes using Eq. \ref{eq:9} based on their salience with respect to the query points. The value of the regularization parameter $\alpha$ (see Eq. \ref{eq:8}) is selected through 10-fold cross-validation. We similarly tune the hyper-parameters $\beta$ to 0.99 and set the number of query points to ten times the number of classes.

After adding salient edges and eliminating non-salient edges through the ranking process, both the adjacency matrix of the modified graph and the node features are passed as input to a two-layer graph convolutional network as described in Section \ref{sec:methods}. 
All of the GCN-based models, including Multi-GCN, are trained for a maximum of 200 iterations, using \textit{Adam} (Adaptive moment estimation extension to stochastic gradient descent -- see \citet{kingma2014adam}) and a learning rate of 0.01. 
Other GCN hyper-parameters are set using the same values reported in \citet{kipf2016semi}.

\begin{table*}[!htb]\centering
	\begin{threeparttable}
		\begin{tabular}{l*{1}{cc}}
			\addlinespace
			\toprule
			
			\multicolumn{3}{c}{\parbox{4.5cm}{\textbf{Predefined train-test splits}   }}\\
			\textbf{Method} & \textbf{Citeseer} & \textbf{Cora}\\
			\midrule
			ManiReg (first view) - \citet{planetoid} &60.1&59.5 \\
			DeepWalk (first view) - \citet{deepwalk} &43.2&67.2 \\
			Planetoid (first view) - \citet{planetoid} &64.7 &75.7\\
			GCN (first view)  & 70.3& 81.5 \\
			GCN (second view) & 50.7& 53.6  \\
			GCN (view union) & 70.7& 80.4\\
			\textbf{Multi-GCN (this paper)} & \textbf{71.3}& \textbf{82.5}  \\
			\midrule 
			\multicolumn{3}{c}{\parbox{4.5cm}{\textbf{Randomized train-test splits}   }}\\
			GCN (first view)  &  67.9$\pm$ 0.5& 80.1$\pm$0.5\\
			GCN (second view) &  53.6$\pm$0.1& 56.9$\pm$0.3\\
			GCN (view union)  & 67.9$\pm$0.3& 78.5$\pm$0.1\\
			\textbf{Multi-GCN (this paper)} & \textbf{70.5$\pm$ 0.2}& \textbf{81.1$\pm$0.2}\\
			\bottomrule
		\end{tabular}
		\caption{\textbf{Classification accuracy on citation networks.} \normalfont{Top panel shows the mean classification accuracy (percentage) for the pre-defined test-train splits as described by \citet{planetoid}. Bottom panel shows the classification accuracy (percentage) and standard error over 10 randomly selected dataset splits of equal size.}\label{tab:results_broader}}
	\end{threeparttable}
\end{table*}

\section{Results}
Experimental results for the three developing-country datasets are shown in Table~\ref{tab:results_random}. Each row in this table indicates the average and standard error of the classification accuracy over 10 randomly drawn train-test splits of the same size for each dataset, constructed as described in Section~\ref{Experiments}. The last row in Table~\ref{tab:results_random} shows the performance of Multi-GCN. In all four datasets, Multi-GCN outperforms existing state-of-the-art benchmarks, with the margin of improvement greatest in the poverty prediction task and smallest in the gender prediction task.

The second set of experimental results, comparing Multi-GCN to recent benchmarks on a more standard node classification task, are shown in Table~\ref{tab:results_broader}. 
In addition to performing a comparison over randomly drawn train-test splits, we also compare the performance of Multi-GCN against a different set of randomized test-train splits, as used in the original tests by  \citet{kipf2016semi}, with an additional validation set of 500 instances used for hyper-parameter tuning. In all cases, we observe improvements in predictive accuracy of Multi-GCN relative to existing approaches.
\section{Discussion}
This paper proposes a new approach to semi-supervised learning on multi-view graphs. Through a series of experiments, we show that this approach improves upon state-of-the-art embedding- and convolution-based algorithms on a variety of prediction tasks related to both poverty research and to node labelling in general. 

Relative to single-view learning algorithms, the main value of the multi-GCN approach is that it incorporates non-redundant information from multiple views into the learning process. Thus, the gains from multi-GCN depend on the prediction task, and the importance of multi-view graph structure to that task. Intuitively, this depends on the mutual information between. This intuition is also supported by a closer look at the results in Table~\ref{tab:results_random}. Here, we observe that while Multi-GCN provides the biggest gains relative to Deepwalk, Node2vec and LINE in the case of product adoption, the gains relative to single-view GCN are more modest. By contrast, the performance gain on the poverty and gender prediction tasks is significantly higher for Multi-GCN, even relative to the other single-view GCN benchmarks. The spy plots in Figures \ref{fig:productadoption_spy}-\ref{fig:gender_spy} help explain this pattern. In particular, we can see that different views in the product adoption setting appear somewhat redundant, whereas for poverty and gender prediction the views appear more independent. 

We believe future work should explore several limitations of the current analysis. In particular, there is much to be learned from a more systematic exploration of the value of additional views, and for different methods for merging views (beyond the subspace learning approach developed in Section~\ref{subsec:methods_subspace}). We are also exploring how graphs with varying degrees of sparsity and a different fraction of labeled nodes can impact the performance of Multi-GCN relative to alternative approaches.

\section{Conclusion}
Graph convolutional networks have recently achieved considerable success in a variety of learning tasks on irregular, graph-structured data. Leveraging insights from spectral graph theory, GCN's are beginning to replicate the success that CNN's have seen on more regular image and text data. For a wide variety of learning tasks relevant to graph-structured data --- in contexts ranging from advertising in online networks to intervening in the spread of a contagious disease --- this is a promising development.

In this paper, we have shown that state-of-the-art GCNs can achieve even greater performance on a variety of classification tasks when the multi-view nature of the underlying network is incorporated into the learning process. While motivated by three applications in global poverty research, the performance gains appear to generalize to other graph-based classification problems. We therefore view Multi-GCN as an important first step in adapting neural network-based approaches to multi-view networks and hope that it  provides a foundation for future work in this space. 

\section{Acknowledgements}
This research was supported by the National Science Foundation Grant under award \#CCF - 1637360 (Algorithms in the Field) and by the Office of Naval Research (Minerva Initiative) under award N00014-17-1-2313.
\clearpage
\bibliographystyle{aaai}
\bibliography{aaairef}

\begin{thebibliography}{}

\bibitem[\protect\citeauthoryear{Blumenstock, Cadamuro, and
  On}{2015}]{blumenstock2015predicting}
Blumenstock, J.; Cadamuro, G.; and On, R.
\newblock 2015.
\newblock Predicting poverty and wealth from mobile phone metadata.
\newblock {\em Science} 350(6264):1073--1076.

\bibitem[\protect\citeauthoryear{Blumenstock}{2014}]{blumenstock2014}
Blumenstock, J.~E.
\newblock 2014.
\newblock Calling for {Better} {Measurement}: {Estimating} an {Individual}’s
  {Wealth} and {Well}-{Being} from {Mobile} {Phone} {Transaction} {Records}.
\newblock In {\em The 20th {ACM} {Conference} on {Knowledge} {Discovery} and
  {Mining} ({KDD} '14), {Workshop} on {Data} {Science} for {Social} {Good}}.

\bibitem[\protect\citeauthoryear{Blumenstock}{2016}]{blumenstock_fighting_2016}
Blumenstock, J.~E.
\newblock 2016.
\newblock Fighting poverty with data.
\newblock {\em Science} 353(6301):753--754.

\bibitem[\protect\citeauthoryear{Bruna \bgroup et al\mbox.\egroup
  }{2013}]{bruna2013spectral}
Bruna, J.; Zaremba, W.; Szlam, A.; and LeCun, Y.
\newblock 2013.
\newblock Spectral networks and locally connected networks on graphs.
\newblock {\em arXiv preprint arXiv:1312.6203}.

\bibitem[\protect\citeauthoryear{Defferrard, Bresson, and
  Vandergheynst}{2016}]{defferrard2016convolutional}
Defferrard, M.; Bresson, X.; and Vandergheynst, P.
\newblock 2016.
\newblock Convolutional neural networks on graphs with fast localized spectral
  filtering.
\newblock In {\em Advances in Neural Information Processing Systems},
  3844--3852.

\bibitem[\protect\citeauthoryear{Dong \bgroup et al\mbox.\egroup
  }{2014}]{dong2014subspace}
Dong, X.; Frossard, P.; Vandergheynst, P.; and Nefedov, N.
\newblock 2014.
\newblock Clustering on multi-layer graphs via subspace analysis on grassmann
  manifolds.
\newblock {\em IEEE Transactions on signal processing} 62(4):905--918.

\bibitem[\protect\citeauthoryear{Farseev \bgroup et al\mbox.\egroup
  }{2017}]{farseev2017cross}
Farseev, A.; Samborskii, I.; Filchenkov, A.; and Chua, T.-S.
\newblock 2017.
\newblock Cross-domain recommendation via clustering on multi-layer graphs.
\newblock In {\em Proceedings of the 40th International ACM SIGIR Conference on
  Research and Development in Information Retrieval},  195--204.
\newblock ACM.

\bibitem[\protect\citeauthoryear{Fatehkia, Kashyap, and Weber}{2018}]{fatehkia}
Fatehkia, M.; Kashyap, R.; and Weber, I.
\newblock 2018.
\newblock Using facebook ad data to track the global digital gender gap.
\newblock {\em World Development} 107:189--209.

\bibitem[\protect\citeauthoryear{Francis, Blumenstock, and
  Robinson}{2017}]{francis2017digital}
Francis, E.; Blumenstock, J.; and Robinson, J.
\newblock 2017.
\newblock Digital credit: A snapshot of the current landscape and open research
  questions.
\newblock {\em CEGA White Paper}.

\bibitem[\protect\citeauthoryear{Frias-Martinez, Frias-Martinez, and
  Oliver}{2010}]{frias2010gender}
Frias-Martinez, V.; Frias-Martinez, E.; and Oliver, N.
\newblock 2010.
\newblock A gender-centric analysis of calling behavior in a developing economy
  using call detail records.
\newblock In {\em AAAI spring symposium: artificial intelligence for
  development}.

\bibitem[\protect\citeauthoryear{Grover and Leskovec}{2016}]{node2vec}
Grover, A., and Leskovec, J.
\newblock 2016.
\newblock Node2vec: Scalable feature learning for networks.
\newblock In {\em Proceedings of the 22Nd ACM SIGKDD International Conference
  on Knowledge Discovery and Data Mining}, KDD '16.
\newblock ACM.

\bibitem[\protect\citeauthoryear{GSMA}{2016}]{gsma_unlocking_2016}
GSMA.
\newblock 2016.
\newblock Unlocking rural coverage: Enablers for commercially sustainable
  mobile network expansion.
\newblock Technical report.

\bibitem[\protect\citeauthoryear{Jean \bgroup et al\mbox.\egroup
  }{2016}]{jean2016combining}
Jean, N.; Burke, M.; Xie, M.; Davis, W.~M.; Lobell, D.~B.; and Ermon, S.
\newblock 2016.
\newblock Combining satellite imagery and machine learning to predict poverty.
\newblock {\em Science} 353(6301).

\bibitem[\protect\citeauthoryear{Khan and Blumenstock}{2016}]{kdd_2016_mrk}
Khan, M.~R., and Blumenstock, J.~E.
\newblock 2016.
\newblock Predictors without borders: Behavioral modeling of product adoption
  in three developing countries.
\newblock In {\em Proceedings of the 22Nd ACM SIGKDD International Conference
  on Knowledge Discovery and Data Mining}, KDD '16.
\newblock ACM.

\bibitem[\protect\citeauthoryear{Kingma and Ba}{2014}]{kingma2014adam}
Kingma, D.~P., and Ba, J.
\newblock 2014.
\newblock Adam: A method for stochastic optimization.
\newblock {\em arXiv preprint arXiv:1412.6980}.

\bibitem[\protect\citeauthoryear{Kipf and Welling}{2017}]{kipf2016semi}
Kipf, T.~N., and Welling, M.
\newblock 2017.
\newblock Semi-supervised classification with graph convolutional networks.
\newblock In {\em International Conference on Learning Representations (ICLR)}.

\bibitem[\protect\citeauthoryear{Liu, Wang, and Chang}{2012}]{liu2012robust}
Liu, W.; Wang, J.; and Chang, S.-F.
\newblock 2012.
\newblock Robust and scalable graph-based semisupervised learning.
\newblock {\em Proceedings of the IEEE} 100(9):2624--2638.

\bibitem[\protect\citeauthoryear{Lu, Bengtsson, and
  Holme}{2012}]{lu2012haitiearthquake}
Lu, X.; Bengtsson, L.; and Holme, P.
\newblock 2012.
\newblock Predictability of population displacement after the 2010 haiti
  earthquake.
\newblock {\em Proceedings of the National Academy of Sciences}
  109(29):11576--11581.

\bibitem[\protect\citeauthoryear{Mikolov \bgroup et al\mbox.\egroup
  }{2013}]{mikolov2013distributed}
Mikolov, T.; Sutskever, I.; Chen, K.; Corrado, G.~S.; and Dean, J.
\newblock 2013.
\newblock Distributed representations of words and phrases and their
  compositionality.
\newblock In {\em Advances in neural information processing systems},
  3111--3119.

\bibitem[\protect\citeauthoryear{Mislove \bgroup et al\mbox.\egroup
  }{2011}]{mislove2011understanding}
Mislove, A.; Lehmann, S.; Ahn, Y.-Y.; Onnela, J.-P.; and Rosenquist, J.~N.
\newblock 2011.
\newblock Understanding the demographics of twitter users.
\newblock {\em ICWSM} 11(5th):25.

\bibitem[\protect\citeauthoryear{Perozzi, Al-Rfou, and Skiena}{2014}]{deepwalk}
Perozzi, B.; Al-Rfou, R.; and Skiena, S.
\newblock 2014.
\newblock Deepwalk: Online learning of social representations.
\newblock In {\em Proceedings of the 20th ACM SIGKDD International Conference
  on Knowledge Discovery and Data Mining}, KDD '14.
\newblock ACM.

\bibitem[\protect\citeauthoryear{Quercia \bgroup et al\mbox.\egroup
  }{2012}]{quercia2012tracking}
Quercia, D.; Ellis, J.; Capra, L.; and Crowcroft, J.
\newblock 2012.
\newblock Tracking gross community happiness from tweets.
\newblock In {\em Proceedings of the ACM 2012 conference on computer supported
  cooperative work},  965--968.
\newblock ACM.

\bibitem[\protect\citeauthoryear{Suri}{2017}]{suri}
Suri, T.
\newblock 2017.
\newblock Mobile money.
\newblock {\em Annual Review of Economics} 9(1):497--520.

\bibitem[\protect\citeauthoryear{Tang \bgroup et al\mbox.\egroup
  }{2015}]{tang2015line}
Tang, J.; Qu, M.; Wang, M.; Zhang, M.; Yan, J.; and Mei, Q.
\newblock 2015.
\newblock Line: Large-scale information network embedding.
\newblock In {\em Proceedings of the 24th International Conference on World
  Wide Web},  1067--1077.
\newblock International World Wide Web Conferences Steering Committee.

\bibitem[\protect\citeauthoryear{Von~Luxburg}{2007}]{von2007tutorial}
Von~Luxburg, U.
\newblock 2007.
\newblock A tutorial on spectral clustering.
\newblock {\em Statistics and computing} 17(4):395--416.

\bibitem[\protect\citeauthoryear{Wesolowski \bgroup et al\mbox.\egroup
  }{2015}]{wesolowski2015dengue}
Wesolowski, A.; Qureshi, T.; Boni, M.~F.; Sunds{\o}y, P.~R.; Johansson, M.~A.;
  Rasheed, S.~B.; Eng{\o}-Monsen, K.; and Buckee, C.~O.
\newblock 2015.
\newblock Impact of human mobility on the emergence of dengue epidemics in
  pakistan.
\newblock {\em Proceedings of the National Academy of Sciences}
  112(38):11887--11892.

\bibitem[\protect\citeauthoryear{Xu, Tao, and Xu}{2013}]{xu2013survey}
Xu, C.; Tao, D.; and Xu, C.
\newblock 2013.
\newblock A survey on multi-view learning.
\newblock {\em arXiv preprint arXiv:1304.5634}.

\bibitem[\protect\citeauthoryear{Yang, Cohen, and
  Salakhutdinov}{2016}]{planetoid}
Yang, Z.; Cohen, W.~W.; and Salakhutdinov, R.
\newblock 2016.
\newblock Revisiting semi-supervised learning with graph embeddings.
\newblock In {\em Proceedings of the 33rd International Conference on
  International Conference on Machine Learning-Volume 48},  40--48.
\newblock JMLR. org.

\bibitem[\protect\citeauthoryear{Zhou \bgroup et al\mbox.\egroup
  }{2004a}]{zhou2004learning}
Zhou, D.; Bousquet, O.; Lal, T.~N.; Weston, J.; and Sch{\"o}lkopf, B.
\newblock 2004a.
\newblock Learning with local and global consistency.
\newblock In {\em Advances in neural information processing systems},
  321--328.

\bibitem[\protect\citeauthoryear{Zhou \bgroup et al\mbox.\egroup
  }{2004b}]{zhou2004ranking}
Zhou, D.; Weston, J.; Gretton, A.; Bousquet, O.; and Sch{\"o}lkopf, B.
\newblock 2004b.
\newblock Ranking on data manifolds.
\newblock In {\em Advances in neural information processing systems},
  169--176.

\bibitem[\protect\citeauthoryear{Zhu}{2005}]{zhu05survey}
Zhu, X.
\newblock 2005.
\newblock Semi-supervised learning literature survey.
\newblock Technical Report 1530, Computer Sciences, University of
  Wisconsin-Madison.

\end{thebibliography}

\end{document}